\documentclass{article}
\usepackage[utf8]{inputenc}
\usepackage[]{algorithm2e}
\usepackage{graphicx}
\usepackage{setspace}
\usepackage[]{amsmath}
\usepackage[]{booktabs}
\usepackage[]{abstract}
\usepackage{longtable}
\usepackage{titlesec}
\usepackage{longtable}
\usepackage{enumitem}
\usepackage{float}
\usepackage[titletoc]{appendix}

\newenvironment{blockquote}{%
  \par%
  \medskip
  \leftskip=4em\rightskip=2em%
  \noindent\ignorespaces}{%
  \par\medskip}
\titleformat*{\section}{\LARGE\bfseries}
\titleformat*{\subsection}{\Large\bfseries}
\titleformat*{\subsubsection}{\large\bfseries}
\titleformat*{\paragraph}{\large\bfseries}
\titleformat*{\subparagraph}{\large\bfseries}

\title{Syntactic and semantic classification of verb arguments using dependency based and rich semantic features}
\author{Francesco Elia\footnote{1243034.elia@studenti.uniroma1.it}\\
Università di Roma ``La Sapienza"}
\date{}
\begin{document}

\maketitle

\begin{abstract}
\noindent Corpus Pattern Analysis (CPA) has been the topic of Semeval 2015 Task 15, aimed at producing a system that can aid lexicographers in their efforts to build a dictionary of meanings for English verbs using the CPA annotation process. CPA parsing is one of the subtasks which this annotation process is made of and it is the focus of this report.  A supervised machine-learning approach has been implemented, in which syntactic features derived from parse trees and semantic features derived from WordNet and word embeddings are used. It is shown that this approach performs well, even with the data sparsity issues that characterize the dataset, and can obtain better results than other system by a margin of about 4\% f-score.
\end{abstract}

\section{Introduction}
\noindent Recent research on Corpus Pattern Analysis, a corpus-driven technique for identifying and assigning meaning to patterns of word usage in text, suggests that it may be useful to build a semantic resource that can be used in several NLP applications. As of now, the main output of CPA is the Pattern Dictionary of English Verbs (\texttt{http://www.pdev.org.uk}), a manually built collection of patterns with entries for each verb in the English language. Task 15 at SemEval 2015 focused on Corpus Pattern Analysis and PDEV, with the aim of producing systems that can automatically build a pattern dictionary for verbs using CPA. To perform this, several stages of processing are needed. The first one, called ``CPA parsing'', is the focus of this report. The CPA parsing task requires a system to identify and classify, from a syntactic and semantic perspective, the relevant arguments for a target verb in a given sentence. The task is similar to Semantic Role Labeling, but single tokens are identified in the dependency parsing paradigm, rather than phrases in the constituency parse tree.
In this report, a system that can perform this task using a learning-based approach is illustrated, by training three maximum entropy classifiers that perform argument identification, syntactic and semantic classification. A rich set of both syntactic and semantic features has been used, showing that they are effective at performing the given task.
The system improves on previous results on the same task, with an f-score increase of almost 4\% on the best performing system.
These results show that, despite the data sparsity that characterizes the dataset provided for this task, a learning-based approach can perform well with the use of descriptive features and that, most likely, this approach would perform even better if more data was available.

\section{CPA parsing}
\indent CPA parsing requires a system to analyze a sentence, extract the verb's main arguments and tag them both syntactically (see Table \ref{syntags} for the syntactic tagset) and semantically (using the CPA ontology\footnote{Available at: http://pdev.org.uk/\#onto}). As an example, let's consider the following sentence, where the verb ``continue'' is the target verb, that is the verb whose arguments have to be extracted:

\begin{blockquote}
European politicians \underline{continue} to plead, sincerely, that Yugoslavia should endure.
\end{blockquote}

\noindent The goal of CPA parsing is to annotate this sentence as shown in Table \ref{cpa-anno}.

\begin{table}[!hbpt]
\centering
\begin{tabular}{@{}l|l|l@{}}
\toprule
\textbf{Token}    & \textbf{Syntactic tag}       & \textbf{Semantic tag}   \\ \midrule
European    &        &            \\
politicians & subj    & Human       \\
continue    & v       & \_          \\
to          & advprep & LexicalItem \\
plead       & acomp   & Activity    \\
,           &        &            \\
sincerely   &        &            \\
,           &        &            \\
that        &        &            \\
Yugoslavia  &        &            \\
should      &        &            \\
endure      &        &            \\
.           &        &           \\ \bottomrule
\end{tabular}
\caption{An example of correctly annotated sentence using CPA syntactic and semantic tag sets}
\label{cpa-anno}
\end{table}

In this task, the target verb is the only information passed to the system, since it is already marked in each sentence in the dataset and thus does not need to be identified. The most important part in solving this task is undoubtedly identifying which tokens in the sentence are actual arguments for the verb, because mistakes at this stage will reflect on all the following steps and will negatively impact the performance of the system. For this reason, similarly to the approach used in \cite{cmills}, I divided the CPA parsing task into three smaller subtasks: argument identification, syntactic classification and semantic classification. Although a single classifier could be used to jointly identify and syntactically classify each token (i.e., by directly assigning a label of ``subj'', ``obj'', etc... when a token is an argument and assigning a label of ``none'' when the token is not) this division allows to separately study which features work best for argument identification and thus improve it without having to worry about the syntactic classification yet. \\
\begin{table}[!hbpt]
\centering
\label{syntags}
\begin{tabular}{@{}l|l@{}}
\toprule
Tag     & Definition                                            \\ \midrule
obj     & Object                                                \\
subj    & Subject                                               \\
advprep & Adverbial preposition or  other Adverbial/Verbal link \\
acomp   & Adverbial or Verb complement                          \\
scomp   & Noun or Adjective complement                          \\
iobj    & Indirect object                                       \\ \bottomrule
\end{tabular}
\caption{Syntactic tagset for CPA parsing}
\end{table}

\begin{figure}[!t]
\centering
\includegraphics[width=0.3\textwidth]{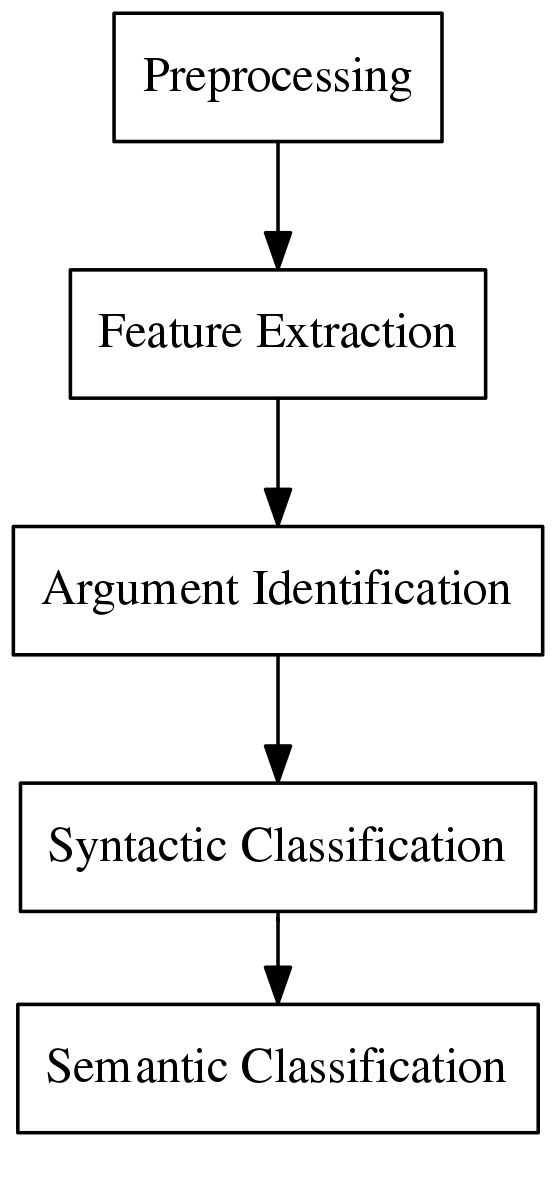}
\caption{Data processing pipeline}
\label{fig-pipeline}
\end{figure}

The three steps of argument identification, syntactic and semantic classification all use a machine-learning approach, implemented with a maximum entropy classifier trained using the Stanford Classifier implementation\footnote{Available at: http://nlp.stanford.edu/software/classifier.shtml}. Maximum entropy models, also known as log-linear or exponential, have been widely used for several NLP tasks, for example question classification, named entity recognition and sentiment analysis; they provide the capability to be trained with several thousands features, which is very common for applications in NLP, and, to a certain extent, they can distringuish between more and less relevant features by automatically assigning higher weights to the former  \cite{stanford-class}. Figure \ref{fig-pipeline} shows the data processing pipeline and how these three modules are arranged with respect to the other components of the system, namely preprocessing and feature extraction.

The preprocessing step, described in detail in Section \ref{sec-preprocessing}, is needed to parse the input sentence and compute syntactic and semantic information that is later used for feature extraction. Several features and their combinations have been tested for each classifier and the best performing ones have been chosen using a greedy hill-climbing algorithm as described in Section \ref{sec-features}. The argument identification, syntactic and semantic classification processes are described in detail in Sections \ref{sec-argid}, \ref{sec-synclass} and \ref{sec-semclass} respectively. 

\subsection{Preprocessing}
\label{sec-preprocessing}
The preprocessing step augments each sentence in the dataset with the linguistic and syntactic information needed to later perform the feature extraction step. All the syntactic information is obtained by using the Stanford CoreNLP pipeline \cite{stanford-corenlp}: in particular, each sentence in the dataset is augmented with POS tags, lemmas, a constituency and a dependency parse tree. \\

\subsection{Features and feature selection}
\label{sec-features}
Despite the system has to work on three different tasks, features have not been created specifically for each one of them: considering the fact that the argument identification and syntactic classification tasks may require very similar features and that these features, despite syntactic in nature, may also benefit the semantic classification task anyway, a unique set of features has been devised and the work of selecting the best performing subset of features for each task is left to the feature selection algorithm. The algorithm starts with an initial, possibly empty, subset of features (provided by the user) and tries one new feature at a time: a new classifier is trained and tested with the new set of features and the new feature is kept if it improves the f-score on the dev set. When there are no more features to be added the algorithm terminates. 
\\
\begin{algorithm}[H]
\label{feature-selection}
 \KwData{S = initial subset of features, A = all the features excluding those in S}
 \KwResult{Subset of features S with the highest f-score}
 bestFscore = 0\;
 \While{A is not empty} {
  feature = sample(A)\;
  A.remove(feature)\;
  fscore = fscore($S \cup \{feature\}$)\;
  \If{$fscore > bestFscore$} {
   S.add(feature)\;
   bestFscore = fscore\;
   }
 }
 return S\;
 \caption{Hill-climbing search for best combination of features}
\end{algorithm}

\bigskip
For compactness reasons, the complete list of features, alogn with a description for each one, is provided in Appendix A.

\subsection{Argument identification}
\label{sec-argid}
Argument identification is performed by a binary classifier that works locally on each token in the sentence and decides whether it is an argument or not. The training set for this classifier is obtained by taking all the tokens in the training set and labeling them with the "argument" class when they have a syntactic label or with "none" otherwise. As an example, how a sample sentence gets annotated by the argument identification classifier:

\begin{table}[!hbpt]
\centering
\begin{tabular}{@{}l|l@{}}
\toprule
\textbf{Token}    & \textbf{Class}   \\ \midrule
Universities & argument\\
continued & verb\\
to & argument\\
languish & argument\\
through & \\
the & \\
eighties & \\
\bottomrule
\end{tabular}
\caption{Output of the argument identification classifier on the example sentence. Tokens that do not represent arguments are labeled with ``none'' but, for simplicity, it is not shown in the table.}
\end{table}

It must be stressed that argument identification is probably the most important step in the whole process, because if an argument is missed or a non-argument is erroneously identified as one, these errors will necessarily propagate through the pipeline rendering the execution of the following steps meaningless.

\subsection{Syntactic classification}
\label{sec-synclass}
This classifier operates on the output of the argument identification step. Tokens identified as arguments are passed to the syntactic classifier, which assigns each of them to one of the 6 possible syntactic classes (subj, obj, iobj, advprep, acomp, scomp). Continuing with the same example sentence, three tokens have been identified as arguments and they have to be classified syntactically:

\begin{table}[!hbpt]
\centering
\begin{tabular}{@{}l|l@{}}
\toprule
\textbf{Token}    & \textbf{Syntactic class}   \\ \midrule
Universities & subj\\
continued & verb\\
to & advprep\\
languish & acomp\\
through & \\
the & \\
eighties & \\
\bottomrule
\end{tabular}
\caption{Output of the syntactic classifier on the example sentence.}
\end{table}

\subsection{Semantic classification}
\label{sec-semclass}
The semantic classifier operates on the output of the previous steps and works in a similar fashion. Each token that has been recognized as an argument gets assigned a semantic label, chosen from the 118 labels present in the training set. Here's how the example sentence that has been considered up to now gets annotated:

\begin{table}[!hbpt]
\centering
\begin{tabular}{@{}l|l|l@{}}
\toprule
\textbf{Token}   & \textbf{Syntactic class} & \textbf{Semantic class}   \\ \midrule
Universities & subj & \textbf{Institution}\\
continued & verb & \textbf{\_}\\
to & advprep & \textbf{LexicalItem}\\
languish & acomp & \textbf{Action}\\
through &  & \\
the & & \\
eighties & & \\
\bottomrule
\end{tabular}
\caption{Output of the semantic classifier on the example sentence, which is now fully annotated.}
\end{table}

\section{Experiments}
\label{sec-experiments}
\subsection{Overview of the dataset}
\label{data-overview}
The performance of the system has been evaluated on the SemEval 2015 Task 15 dataset.
The training set is made of 3249 sentences for 21 different verbs, while the test set contains 1280 sentences for 7 verbs. It must be noted that the verbs contained in the training set are different than those contained in the test set, in order to encourage researchers to work on systems that could generalize to different verbs than those they have been trained on. Table \ref{verb-stats} shows the number of sentences for each verb in the training and test set.

\begin{table}[!hbpt]
\centering
\begin{tabular}{@{}ll|ll@{}}
\toprule
\textbf{Verb}    & \textbf{\# of sentences}  & \textbf{Verb}    & \textbf{\# of sentences} \\ \midrule
allow & 150 & sabotage & 77 \\
crave & 75 & recall & 263 \\
launch* & 207 & claim & 212 \\
propose & 169 & applaud & 198 \\
execute* & 213 & veto & 123 \\
pray* & 180 & plan & 130 \\
announce & 228 & account & 155 \\
battle* & 190 & \underline{undertake} & 228 \\
plead & 205 & \underline{crush} & 170 \\
abandon & 172 & \underline{operate} & 140 \\
answer & 171 & \underline{apprehend} & 123 \\
abort & 60 & \underline{appreciate} & 215 \\
squeal & 20 & \underline{continue} & 203 \\
disable & 51 & \underline{decline} & 201 \\
\bottomrule
\end{tabular}
\caption{Number of sentences for each verb in the dataset. Underlined verbs are part of the test set. Verbs marked with * have been chosen for the dev set.}
\label{verb-stats}
\end{table}

\begin{figure}[!hbpt]
  \centering
    \includegraphics[width=0.8\textwidth]{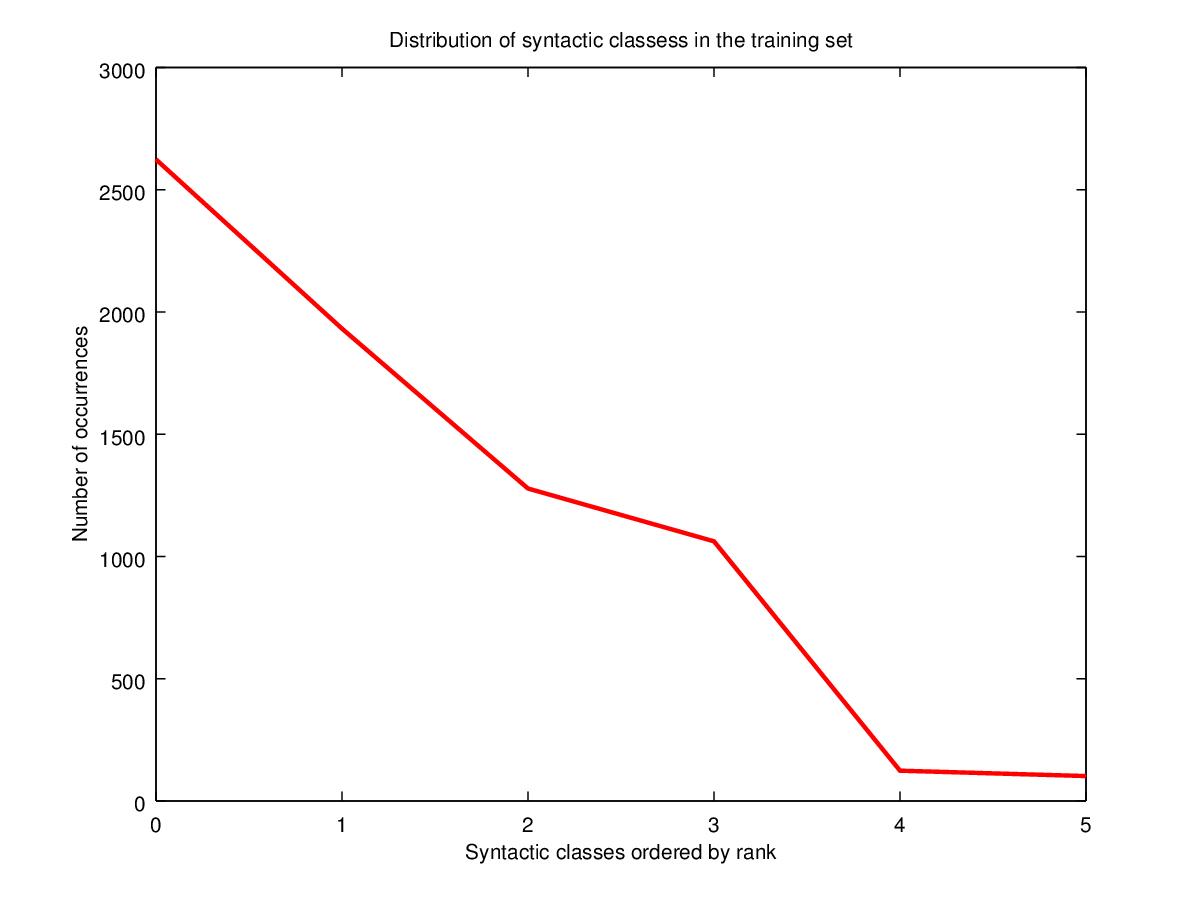}
    \label{fig-syntactic-distribution}
    \caption{Distribution of syntactic classes in the training set.}
\end{figure}

The training set has a total of 7122 tokens with a syntactic and semantic label. Figure \ref{fig-syntactic-distribution} shows how syntactic classes are distributed in the training set: the most frequent class (rank 0) is \emph{subj} followed by \emph{obj}, \emph{acomp}, \emph{advprep}, \emph{scomp} and finally \emph{iobj}. The last two classes (\emph{iobj} and \emph{scomp}) are an order of magnitude less frequent than all the other ones, but considering the fact that they do not appear in the test set, and thus do not affect the performance of the system, the frequency of the other classes is pretty balanced.

\begin{figure}[!hbt]
  \centering
    \includegraphics[width=0.8\textwidth]{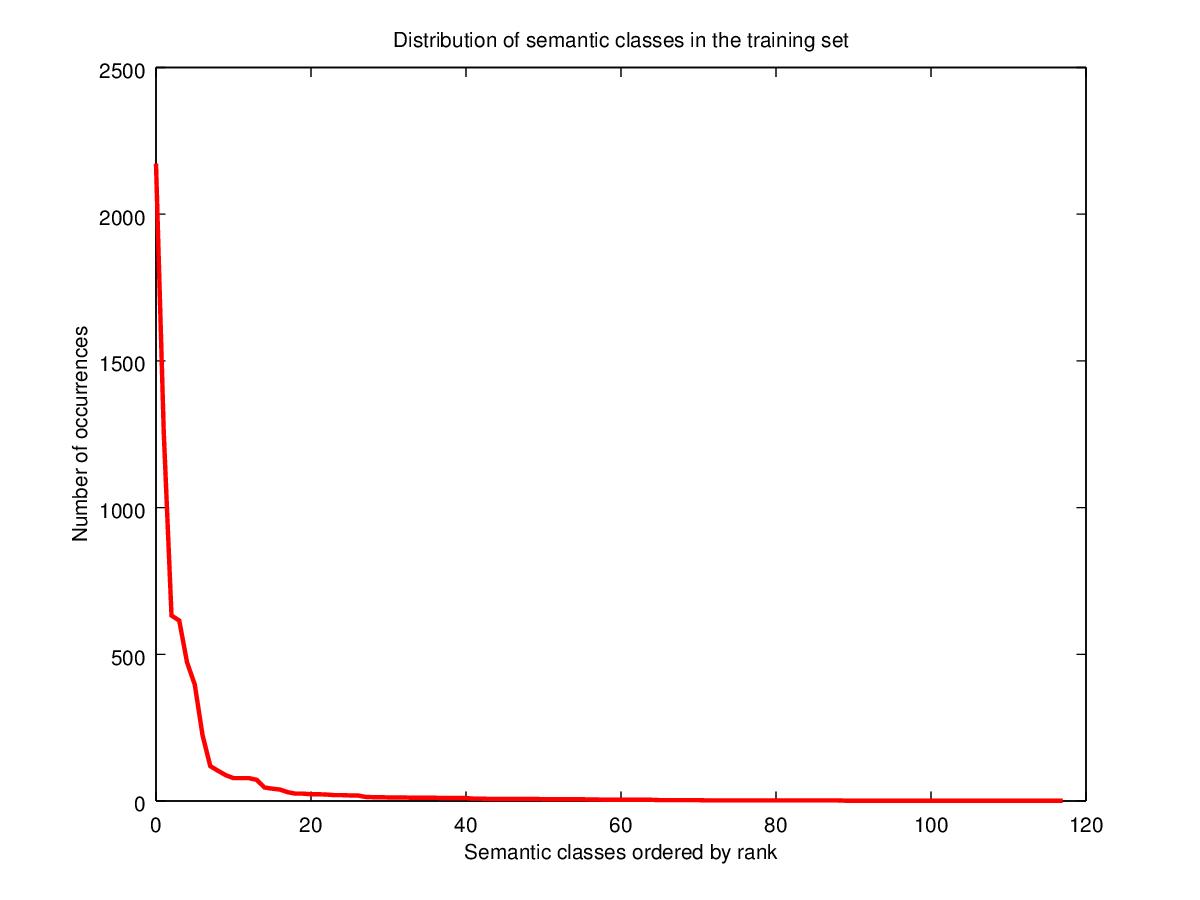}
    \caption{Distribution of semantic classes in the training set. As can be seen, the frequency of a class is roughly inversely proportial to its rank, leading to a very skewed distribution where the top 3 most frequent classes constitute more than half of the training set.}
    \label{fig-semantic-distribution}
\end{figure}

Unfortunately, as can be seen from Figure \ref{fig-semantic-distribution}, the distribution of semantic classes is instead very skewed, with the first three most common classes (i.e., Human, LexicalItem, Action) accounting for 57\% of the total number of examples. Out of the 118 semantic classes that appear in the training set, 77 of them do so less than 10 times and this is potentially an issue for the semantic classification task, that will be shown to be the less performing part of the system. 

\subsection{Metrics}
\label{metrics}
The metric used for the evaluation is the average f-score of the system over each verb in the test set:
$$\text{F1}_{verb} = \frac{2 \cdot  \text{Precision}_{verb} \cdot \text{Recall}_{verb}}{\text{Precision}_{verb} + \text{Recall}_{verb}}$$

$$\text{Score} = \frac{\sum_{verb \in V}\text{F1}_{verb}}{|V|}$$

\noindent Where $V$ is the set of verbs contained in the test set. Precision and recall are calculated as follows:

$$\text{Precision} = \frac{\text{Correct tags}}{\text{Retrieved tags}}$$
$$\text{Recall} = \frac{\text{Correct tags}}{\text{Reference tags}}$$

A tag is considered correct when it is placed on the exact same token and matches the annotation in the gold standard. For this reason, a good performance of the argument identification module is essential for good final results, seeing as errors at this stage will be reflected on all the following ones: if arguments are not identified this will impact recall for syntactic and semantic classification and if tokens are erroneously identified as arguments this will impact their precision, no matter what syntactic/semantic class is assigned to those tokens.

\subsection{Results of feature selection}
\label{sec-feature-selection}
Applying the feature selection algorithm results in three subsets of features, one for each of the three classifiers used in the pipeline, as shown in the following table.
\begin{longtable}{lccc} 
\toprule
Feature name & ArgId & SynClass & SemClass \\ \midrule \midrule
\endhead
\textsc{\scriptsize{TokenLemma}} & \textbullet  & \textbullet  &  \textbullet \\
\textsc{\scriptsize{TokenWord}} &   &   &   \\
\textsc{\scriptsize{TokenPos}} &  \textbullet &   &  \textbullet \\
\textsc{\scriptsize{LemmasAroundToken}} &   &   & \textbullet   \\
\textsc{\scriptsize{WordsAroundToken}} &   &   &   \\
\textsc{\scriptsize{PosAroundToken}} &   &   &   \\
\textsc{\scriptsize{TokenIsVerb}} &   &   &   \\
\textsc{\scriptsize{TokenIsPrepositionOfVerb}} & \textbullet  &   &   \\
\textsc{\scriptsize{TokenPhraseType}} & \textbullet  &   &   \\
\textsc{\scriptsize{TokenPhraseStructure}} &   & \textbullet  &   \\
\textsc{\scriptsize{TokenIsSubjOrObj}} &   & \textbullet  &   \\
\textsc{\scriptsize{TokenIsVerbChild}} & \textbullet  & \textbullet  &   \\
\textsc{\scriptsize{TokenIsCapitalized}} &   & \textbullet  & \textbullet  \\
\textsc{\scriptsize{TokenContainsDigit}} &   &   & \textbullet  \\
\textsc{\scriptsize{TokenIsUppercase}} &   &   &  \textbullet \\
\textsc{\scriptsize{TokenRelFromVerb}} & \textbullet  & \textbullet  &   \\
\textsc{\scriptsize{TokenIsUniqueSubjOrObj}} &   &   &   \\
\textsc{\scriptsize{VerbLemma}} &   &   &   \\
\textsc{\scriptsize{VerbPos}} &   & \textbullet  &   \\
\textsc{\scriptsize{LemmasAroundVerb}} &   &  \textbullet &   \\
\textsc{\scriptsize{PosAroundVerb}} &   & \textbullet  &   \\
\textsc{\scriptsize{VerbVoice}} &   & \textbullet  &   \\
\textsc{\scriptsize{VerbPosition}} &   & \textbullet  &   \\
\textsc{\scriptsize{IsVerbPrepositional}} &   &   &   \\
\textsc{\scriptsize{VerbBy}} &   &   &   \\
\textsc{\scriptsize{VerbPhraseStructure}} &   & \textbullet  &   \\
\textsc{\scriptsize{VerbIsRoot}} &   & \textbullet  &   \\
\textsc{\scriptsize{VerbHasNsubj}} & \textbullet  & \textbullet   &   \\
\textsc{\scriptsize{VerbHasNsubjpass}} & \textbullet   & \textbullet   &   \\
\textsc{\scriptsize{VerbHasDobj}} & \textbullet  & \textbullet  &   \\
\textsc{\scriptsize{VerbHasIobj}} &   &   &   \\
\textsc{\scriptsize{VerbHasCcomp}} &   & \textbullet  &   \\
\textsc{\scriptsize{VerbHasAcomp}} &   &   &   \\
\textsc{\scriptsize{VerbHasXcomp}} & \textbullet  &   &   \\
\textsc{\scriptsize{VerbParentLemma}} &   & \textbullet  & \textbullet  \\
\textsc{\scriptsize{VerbParentPos}} &   &   &   \\
\textsc{\scriptsize{VerbFirstVpParentLemma}} & \textbullet  &   &   \\
\textsc{\scriptsize{RelVerbParentToVerb}} & \textbullet  &   &   \\
\textsc{\scriptsize{TokenDirDpathFromVerb}} & \textbullet  & \textbullet  & \textbullet  \\
\textsc{\scriptsize{TokenDirDpathFromVerbWithLemma}} &   & \textbullet  &  \textbullet \\
\textsc{\scriptsize{TokenDirDpathFromVerbWithPos}} & \textbullet  &   &   \\
\textsc{\scriptsize{TokenUndDpathFromVerb}} & \textbullet  & \textbullet  &   \\
\textsc{\scriptsize{TokenUndDpathFromVerbWithLemma}} &   &   &  \textbullet \\
\textsc{\scriptsize{TokenUndDpathFromVerbWithPos}} & \textbullet  & \textbullet  &   \\
\textsc{\scriptsize{DirDpathVerbVpParentToVerb}} & \textbullet  &   &   \\
\textsc{\scriptsize{DirDpathVerbVpParentToVerbWithLemma}} & \textbullet  & \textbullet   &   \\
\textsc{\scriptsize{DirDpathVerbVpParentToVerbWithPos}} & \textbullet  & \textbullet  &   \\
\textsc{\scriptsize{TokenDirDpathFromVerbVpParent}} & \textbullet  & \textbullet  &   \\
\textsc{\scriptsize{TokenDirDpathFromVerbVpParentWithLemma}} & \textbullet  & \textbullet  &   \\
\textsc{\scriptsize{TokenDirDpathFromVerbVpParentWithPos}} &   & \textbullet  &   \\
\textsc{\scriptsize{TokenParentDirDpathFromVerb}} & \textbullet  & \textbullet  &   \\
\textsc{\scriptsize{TokenParentDirDpathFromVerbWithLemma}} & \textbullet  &   &   \\
\textsc{\scriptsize{TokenParentDirDpathFromVerbWithPos}} & \textbullet  & \textbullet  &   \\
\textsc{\scriptsize{TokenDirDpathFromVerbParent}} & \textbullet  &   &   \\
\textsc{\scriptsize{TokenDirDpathFromVerbParentWithLemma}} &   &   &   \\
\textsc{\scriptsize{TokenDirDpathFromVerbParentWithPos}} &   & \textbullet  &   \\
\textsc{\scriptsize{TokenCpathFromVerb}} & \textbullet  & \textbullet  &   \\
\textsc{\scriptsize{TokenCpathFromVerbParent}} & \textbullet  & \textbullet   &   \\
\textsc{\scriptsize{VerbHypernymsMcs}} & \textbullet  &   & \textbullet  \\
\textsc{\scriptsize{HypernymsMcs}} &   &   & \textbullet  \\
\textsc{\scriptsize{HypernymsDisambiguated}} &   &   &   \\
\textsc{\scriptsize{TokenSimilarWords}} &   &   & \textbullet  \\
\textsc{\scriptsize{VerbSimilarWords}} &   &   &  \textbullet \\
\textsc{\scriptsize{TokenMostSimilarLabels}} &   &   &  \textbullet \\
\textsc{\scriptsize{VerbPreps}} & \textbullet & \textbullet & \\
\textsc{\scriptsize{VerbDistance}} & \textbullet & & \\
\textsc{\scriptsize{DepDepthDifference}} & \textbullet &  \textbullet & \\
\textsc{\scriptsize{ConDepthDifference}} & \textbullet & & \\
\textsc{\scriptsize{DepPathToVerbLength}} & \textbullet & & \\
\textsc{\scriptsize{ConPathToVerbLength}} & & \textbullet & \\
\bottomrule
\end{longtable}

The two tasks of argument identification and syntactic classification share several features: more precisely, most of the features that are useful for argument identification are also useful in classification while this is not true the other way around. A quick look at the section containing \textsc{Verb*} features, shows that most of these are only used in classification. This is expected, since, for example, a combination of features like \textsc{VerbPosition} and \textsc{VerbVoice} is very useful in discriminating between a subject (which usually appears before/after the verb if the sentence is in active/passive form) and an object, which behaves inversely, but not very discriminative for argument identification in general. As could be expected, the \textsc{VerbLemma} feature does not improve the performance of any classifier, because the test/dev sets use sentences with different verbs than those seen during training. For this reason, many features derived from word embeddings (e.g., \textsc{VerbSimilarWords}, \textsc{TokenSimilarWords}) proved to be really useful for semantic classification, allowing the classifier to ``understand'' when a new verb is similar to one it has already seen, thus better generalizing to new unseen verbs instead of treating them as completely unrelated to what it has seen during training.\\

\begin{table}[!hbp]
\centering
\begin{tabular}{@{}lccc@{}}
\toprule
\textbf{Classifier}    & \textbf{Precision}  & \textbf{Recall}    & \textbf{Fscore} \\ \midrule
Argument identification & 0.862 & 0.751 & 0.802 \\
Syntactic classification & 0.945 & --- & --- \\
Semantic classification & 0.695 & --- & --- \\
\bottomrule
\end{tabular}
\caption{The performance of the three models on the test set, with the best subset of features resulting from the application of the feature selection algorithm.}
\label{tab-bestfeatures}
\end{table}

Table \ref{tab-bestfeatures} shows the performance of the classifiers with the best subsets of features obtained by the feature selection algorithm. Recall and f-score are only shown for argument identification, because, as previously mentioned, recall of syntactic and semantic classification is a direct result of that of argument identification and is not considered during the feature selection process since the three steps are optimized independently from each other.

\section{Results}
\label{sec-results}
The performance of the system has been compared to the three other systems that participated at SemEval 2015 task 15 and to that of the baseline provided by the organizers of the task. The following results are obtained using the best performing features for each classifier as detailed in Table \ref{tab-bestfeatures}. The performance of each system, including mine and the baseline, is shown in Table \ref{tab-systems}.

\begin{table}[!hbp]
\centering
\begin{tabular}{@{}l|l@{}}
\toprule
\textbf{System}    &  \textbf{F-score} \\ \midrule
My system & \textbf{0.661}\\
baseline & 0.624\\
FANTASY & 0.589\\
BLCUNLP & 0.530\\
CMILLS & 0.516\\
\bottomrule
\end{tabular}
\caption{Comparison of the scores of the CPA parsing systems.}
\label{tab-systems}
\end{table}

As previously noted, my system improves on the baseline with an increment of almost 4\% f-score and on FANTASY, the best performing system submitted by participants, with an increment of around 7\%. This final score, as explained in Section \ref{metrics}, is calculated as the average f-score for all the verbs in the test set, so a detailed breakdown of precision, recall and f-score for syntactic and semantic classification for each of them is shown in Table \ref{tab-breakdown}.

\begin{table}[!hbp]
\centering
\resizebox{\textwidth}{!}{%
\begin{tabular}{lccccccccc}
\toprule
\textbf{Verb} & \multicolumn{3}{c}{\textbf{Syntactic stats}} & \multicolumn{3}{c}{\textbf{Semantic stats}} & \multicolumn{3}{c}{\textbf{Average stats}} \\
 & precision & recall & fscore & precision & recall & fscore & precision & recall & fscore \\
 \midrule
crush & 0.836      & 0.729 & 0.779 & 0.484  & 0.436 & 0.459 & 0.657 & 0.582 & 0.617 \\
continue & 0.92    & 0.84  & 0.878 & 0.634  & 0.578 & 0.605 & 0.777 & 0.709 & 0.741 \\
operate & 0.788    & 0.532 & 0.635 & 0.327  & 0.21  & 0.256 & 0.563 & 0.371 & 0.447 \\
decline & 0.898    & 0.838 & 0.867 & 0.626 & 0.578 & 0.601 & 0.763 & 0.708 & 0.734 \\
undertake & 0.754  & 0.717 & 0.735 & 0.585  & 0.546 & 0.565 &  0.67 & 0.632 & 0.65 \\
apprehend & 0.825  & 0.722 & 0.77  & 0.745  & 0.634 & 0.685 & 0.786 & 0.678 & 0.728 \\
appreciate & 0.908 & 0.713 & 0.798 & 0.716  & 0.559 & 0.628 & 0.812 & 0.636 & 0.713 \\
\midrule
\textbf{AVERAGE} & \textbf{0.847} & \textbf{0.727} & \textbf{0.78} & \textbf{0.588} & \textbf{0.506} & \textbf{0.543} & \textbf{0.718} & \textbf{0.617} & \textbf{0.661} \\
\bottomrule
\end{tabular}
}
\caption{Detailed scores for each verb in the test set, obtained by using the scorer for the SemEval task available on the official website}
\label{tab-breakdown}
\end{table}

Syntactic classification is definitely the best performing part of the system and analyzing in detail the contribution of precision and recall to the final syntactic score, it can be seen that precision is much higher, which is usually a desirable property in a system that aims at minimizing the work that human annotaors have to do.
Semantic classification has much lower scores, both in terms of precision and recall. As explained in Section \ref{sec-experiments}, semantic classes in the training set follow a very skewed distribution with common classes such as \emph{Human} appearing for 30\% of the tokens. To show how the frequency of a semantic class impacts the f-score that the classifier ends up having, a plot of the f-score of each class with respect to its frequency in the training set is shown in Appendix B. It is evident that the there is a strong correlation between the number of examples and the f-score for each semantic class and, indeed, the best performing classes are the most frequent in the training set.

\begin{table}[!hbp]
\centering
\resizebox{\textwidth}{!}{%
\begin{tabular}{lcccccc}
\toprule
Category & \#Gold & CMILLS & FANTASY & BLCUNLP & baseline & My system\\ \midrule
subj & 1,008 & 0.564 & 0.694 & 0.739 & \textbf{0.815} & 0.785 \\
obj & 777 & 0.659 & {0.792} & 0.777 & 0.783 & \textbf{0.817} \\
Human & 580 & 0.593 & \textbf{0.770} & 0.691 & 0.724 & 0.726 \\
Activity & 438 &  0.450 & 0.479 & 0.393 & 0.408 & \textbf{0.571}  \\
acomp & 308 & 0.545 & 0.418 & 0.702 & \textbf{0.729} & 0.705  \\
LexicalItem & 303 & 0.668 & \textbf{0.830} &  0.771 & 0.811 & 0.766  \\
advprep & 289 & 0.621 & 0.517 & 0.736 & \textbf{0.845} & 0.817  \\
State  Of  Affairs & 192 & 0.410 & 0.276 & 0.373 & 0.211 & \textbf{0.529}  \\
Institution & 182 & 0.441 & \textbf{0.531} & {0.483} & 0.461 & 0.512  \\
Action & 115 & 0.421 & \textbf{0.594} & 0.526 & 0.506 & 0.372 \\
\bottomrule
\end{tabular}
}
\caption{F-scores for the top 10 most frequent classes, compared to those obtained by the other systems.}
\label{tab-top10}
\end{table}

Table \ref{tab-top10} shows the performance of the systems on the top 10 most frequent classes, not distinguishing between syntactic and semantic ones. It can be seen that the baseline is still strong at predicting syntactic arguments using a rule-based approach, but gets beaten in every semantic class. The FANTASY system, which was the best performing one in this task, is not accompained by an article, so it is not possible to analyze their results in depth.

\section{Conclusions}
\label{sec-conclusions}
The CPA parsing task, which has been discussed in this report, is the first step in the CPA annotation process and consists in identifying and classifying, from a syntactic and semantic perspective, the relevant arguments of a target verb in a sentence so that patterns can subsequently be extracted for each verb and clustered together according to their similarity. Syntactic classification had to be performed on a small syntactic tagset comprising the most common syntactic functions (subject, object, indirect object, adjective complement, adverbial complement) while semantic classification used the CPA ontology, which contains around 250 hierarchically organized semantic classes.\\
CPA parsing proved to be a difficult task, as none of the systems that participated at SemEval 2015 Task 15 managed to beat the rule-based baseline that was provided by the task organizers, with the best performance being very close but still 3.5\% lower than that of the baseline. The approach described in this report outperforms these systems and slightly improves on the baselines results as well, with an increase of almost 4\% in terms of f-score. The system uses a pipeline of three maximum entropy classifiers, which, in turn, first identify the relevant verb arguments and then classify them syntactically and semantically. Among these three steps the argument identification one is arguably the most important in the whole process, as errors at this stage will propagate to all the following steps and its performance is an upper bound to that of the whole system. In the proposed system, this stage performs well, with an f-score of 80\%. Syntactic classification, however, is the best performing step in the pipeline: this is probably due to the fact that, given the way the data is structured, the training set contains a high and balanced number of examples for each syntactic class, resulting in easier training. While performance on the syntactic level can be considered good, that of the semantic layer is, unfortunately, very far from those of both argument identification and syntactic classification. There are multiple reasons for this to happen, but, most likely, this is due to the data sparsity problem that characterizes semantic classes in the training set and that has been previously discussed in Section \ref{data-overview}. \\
These three classifiers are trained with several kind of features, some inspired from previous work on Semantic Role Labeling and CPA parsing, while others are novel features or variations on known features designed by me during my work on this task. Given the high number of features, manually testing the effectiveness of each one is a hard task, so a simple feature selection algorithm has been employed in order to select the best subset for each of the three models. Despite this, running the feature selection algorithm is still computationally expensive and I have been limited in the number of iterations I could actually execute. I believe that better feature subsets can be found and also that experimenting with other feature selection techniques could lead to improvements in the performance of the system. \\
In conclusion, I presented a learning-based approach to CPA parsing which showed to perform better than previous attempts, thanks to descriptive features tailored to each of the subtasks that CPA parsing is made of. While most of previous CPA parsing systems either used rule-based approaches or relatively simple features for semantic classification (for example named entities) this work shows how dependency parse trees and WordNet and word embeddings can be used to derive a useful set of features to improves the performance of syntactic and semantic classification respectively.

\clearpage
\renewcommand\thesection{}
\section{Appendices}
  \section{A List of features}
  \begin{enumerate}
  \item \textsc{\small{TokenLemma}} The lemma of the token.

\item \textsc{\small{TokenWord}} The word form of the token.

\item \textsc{\small{TokenPos}} The POS tag of the token.

\item \textsc{\small{LemmasAroundToken}} Lemmas of the tokens immediately before and after the current one. If the token is at the start or at the end of the sentence, special \texttt{<s>} or \texttt{</s>} symbols are used to represent the previous or next token respectively.

\item \textsc{\small{WordsAroundToken}} Words immediately before and after the current token. If the token is at the start or at the end of the sentence, special \texttt{<s>} or \texttt{</s>} symbols are used to represent the previous or next token respectively.

\item \textsc{\small{PosAroundToken}} POS tags of the tokens immediately before and after the current one. If the token is at the start or at the end of the sentence, special \texttt{<s>} or \texttt{</s>} symbols are used to represent the previous or next token respectively.

\item \textsc{\small{TokenIsVerb}} Whether the token is a verb. A token is considered a verb if its POS tag starts with \texttt{V}.

\item \textsc{\small{TokenIsPrepositionOfVerb}} Whether the token is a prepositional dependency of the target verb, that is whether the target verb node in the dependency parse tree has a \emph{prep} dependency relation with the token as child. To compute this feature CoreNLP basic dependencies annotations are used (as opposed to the collapsed dependencies annotations that are used to compute the other features), because when using the other types of dependencies prepositions are collapsed into the edges and they don't have a corresponding node in the tree.

\item \textsc{\small{TokenPhraseType}} The phrase type the token belongs to according to the constituency parse tree. Formally, the value of the first non pre-terminal ancestor of the token node (the pre-terminal node contains POS tag annotations); in the case of the ``Universities'' token the value is \texttt{token\_phrase\_type=NP}.

\item \textsc{\small{TokenPhraseStructure}} This feature represents the structure of the subtree the token belongs to. As for the \textsc{\small{TokenPhraseType}} feature, the first non pre-terminal ancestor of the token node is identified, and then all its children are listed in left to right order; as for the example token the value is \texttt{token\_phrase\_strcture=NP->NNS}.

\item \textsc{\small{TokenIsSubjOrObj}} Whether the token has a parent with a \emph{nsubj}, \emph{nsubjpass} or \emph{dobj} relation in the dependency parse tree.

\item \textsc{\small{TokenIsVerbChild}} Whether the token is a child of the target verb in the dependency parse tree.

\item \textsc{\small{TokenIsCapitalized}} Whether the token is capitalized. This feature should be most useful for semantic classification, in recognizing many proper names that appear in the dataset.

\item \textsc{\small{TokenContainsDigit}} Whether the token contains digits. This feature should be most useful for semantic classification, in recognizing many instances of the \emph{Numerical Value} semantic class.

\item \textsc{\small{TokenIsUppercase}} Whether the token is completely uppercase. This feature should be most useful for semantic classification, since, for example, many instances of the \emph{Business Enterprise} class are written in all uppercase characters.

\item \textsc{\small{TokenRelFromVerb}} The direct relation in the dependency parse tree from the target verb to token (if any).

\item \textsc{\small{TokenIsUniqueSubjOrObj}} Whether or not the current token is the only one in the sentence to have a parent with a \emph{nsubj}, \emph{nsubjpass} and \emph{dobj} relation. More specifically, for each of these relations $R$ the output is a boolean feature (so this actually adds three features at most) that takes its value according to the following rules:
\begin{itemize}
\item if the token doesn't have a parent with relation $R$, this feature is not added;
\item if the token has a parent with relation $R$ but there exists, in the dependency parse tree, two other tokens linked by $R$, the feature is set to false;
\item otherwise it is set to true.
\end{itemize}

As for the example, the features for the \emph{nsubjpass} and \emph{dobj} dependencies will not be set, since the ``Universities'' token doesn't have such relations, and the feature for \emph{nsubj} will be set to \texttt{is\_unique\_nsubj=true} because it is the only token in the whole dependency parse tree to have a parent linked by the \emph{nsubj} relation.

\end{enumerate}
\noindent Features computed on the target verb:
\begin{enumerate}
\setcounter{enumi}{17}
\item \textsc{\small{VerbLemma}} Lemma of the target verb.

\item \textsc{\small{VerbPos}} POS tag of the target verb.

\item \textsc{\small{LemmasAroundVerb}} Lemmas of the tokens immediately before and after the target verb. If the target verb is at the start or at the end of the sentence, special \texttt{<s>} or \texttt{</s>} symbols are used to represent the previous or next token respectively.

\item \textsc{\small{PosAroundVerb}} POS tags of the tokens immediately before and after the verb. If the verb is at the start or at the end of the sentence, special \texttt{<s>} or \texttt{</s>} symbols are used to represent the previous or next token respectively.

\item \textsc{\small{VerbVoice}} Whether the target verb has active or passive voice, determined through the use of dependencies: if the verb has any child with a relation of \emph{nsubjpass}, \emph{csubjpass}, \emph{auxpass} or \emph{agent} the voice is set to passive, otherwise it is set to active.

\item \textsc{\small{VerbPosition}} Whether the current token is before or after the target verb.

\item \textsc{\small{IsVerbPrepositional}} Whether the target verb belongs to a list of prepositional verbs. The list contains verbs that commonly take prepositional arguments like ``continue'' (\emph{continue to}), ``thank'' (\emph{thank for}), etc... and it is shown in Appendix A.

\item \textsc{\small{VerbBy}} Whether the target verb has a \emph{prep\_by} dependency. If this happens, the dependant is probably a passive subject.

\item \textsc{\small{VerbPhraseStructure}} The same as the \textsc{\small{ArgumentPhraseStructure}} feature, but computed for the target verb token.\\ In the example: \texttt{verb\_phrase\_structure=VP->VBD-S}

\item \textsc{\small{VerbIsRoot}} Whether the target verb is the root of the sentence in the dependency parse tree.

\item \textsc{\small{VerbHasNsubj}} Whether the target verb has a \emph{nsubj} dependency.

\item \textsc{\small{VerbHasNsubjpass}} Whether the target verb has a \emph{nsubjpass}
dependency.

\item \textsc{\small{VerbHasDobj}} Whether the target verb has a \emph{dobj} dependency.

\item \textsc{\small{VerbHasIobj}} Whether the target verb has a \emph{iobj} dependency.

\item \textsc{\small{VerbHasCcomp}} Whether the target verb has a \emph{ccomp} dependency.

\item \textsc{\small{VerbHasAcomp}} Whether the target verb has a \emph{acomp} dependency.

\item \textsc{\small{VerbHasXcomp}} Whether the target verb has a \emph{xcomp} dependency.

\item \textsc{\small{VerbParentLemma}} Lemma of the parent of the target verb in the dependency parse tree (if any).

\item \textsc{\small{VerbParentPos}} The POS tag of the parent of the target verb in the dependency parse tree.

\item \textsc{\small{VerbFirstVpParentLemma}} Lemma of the first ancestor of the target verb that is also a verb: if the parent of the target verb is a verb then this feature is the same as \textsc{\small{VerbParentLemma}}.

\item \textsc{\small{RelVerbParentToVerb}} The relation from the target verb parent to the target verb in the dependency parse tree (if any).

\end{enumerate}

\noindent The following features are based on paths between nodes in the dependency parse tree. Six types of paths are considered and each path can be augments with lemmas or POS tags of the traversed nodes, giving rise to a total of 18 feature types based on dependency paths.
\begin{enumerate}[topsep=6pt,itemsep=-1ex,partopsep=1ex,parsep=1ex]
\setcounter{enumi}{38}

\item \textsc{\small{TokenDirPathFromVerb}}
\item \textsc{\small{TokenDirPathFromVerbWithLemma}}
\item \textsc{\small{TokenDirPathFromVerbWithPos}}\\
The shortest directed path in the dependency parse tree from the target verb to the current token. The path is the list of dependency relations that are traversed to reach the candidate token from the verb and it can be augmented with lemmas or POS tags of the traversed nodes. In the example sentence shown before these three features will have the following values:
\begin{itemize}[topsep=6pt,itemsep=-1ex,partopsep=1ex,parsep=1ex]
\item \texttt{token\_dir\_dpath\_from\_verb=nsubj}
\item \texttt{token\_dir\_dpath\_from\_verb\_with\_lemma=nsubj-continue}
\item \texttt{token\_dir\_dpath\_from\_verb\_with\_pos=nsubj-VBD}
\end{itemize}
\noindent If such a path does not exist this feature is not set.\\

\item \textsc{\small{TokenUndDpathFromVerb}}
\item \textsc{\small{TokenUndDpathFromVerbWithLemma}}
\item \textsc{\small{TokenUndDpathFromVerbWithPos}}\\
The shortest undirected path between the target verb and the current token in the dependency parse tree, augmented with lemmas and POS tags of the traversed nodes similarity to the previous three features. This path is guaranteed to exist, so this feature is always set.\\

\item \textsc{\small{DirDpathVerbVpParentToVerb}}
\item \textsc{\small{DirDpathVerbVpParentToVerb}}
\item \textsc{\small{DirDpathVerbVpParentToVerb}}\\
The shortest directed path from the first VP ancestor of the target verb to the target verb.\\

\item \textsc{\small{TokenDirDpathFromVerbVpParent}}
\item \textsc{\small{TokenDirDpathFromVerbVpParentWithLemma}}
\item \textsc{\small{TokenDirDpathFromVerbVpParentWithPos}}\\
The shortest directed path from the first VP ancestor of the target verb to the current token. \\

\item \textsc{\small{TokenParentDirDpathFromVerb}}
\item \textsc{\small{TokenParentDirDpathFromVerbWithLemma}}
\item \textsc{\small{TokenParentDirDpathFromVerbWithPos}}\\
The shortest directed path from the target verb to the token parent.\\

\item \textsc{\small{TokenDirDpathFromVerbParent}}
\item \textsc{\small{TokenDirDpathFromVerbParentWithLemma}}
\item \textsc{\small{TokenDirDpathFromVerbParentWithPos}}\\
The shortest directed path from the verb parent in the dependency parse tree to the current token.\\
\end{enumerate}

\noindent Two features based on paths in the constituency parse tree:
\begin{enumerate}[topsep=6pt,itemsep=-1ex,partopsep=1ex,parsep=1ex]
\setcounter{enumi}{56}

\item \textsc{\small{TokenCpathFromVerb}} The path in the constituency parse tree from the target verb to the current token. 
\item \textsc{\small{TokenCpathFromVerbParent}} The path in the constituency parse tree from the parent of the target verb to the current token. 

\end{enumerate}

\noindent Purely semantic features:
\begin{enumerate}
\setcounter{enumi}{58}
\item \textsc{\small{HypernymsMCS}} All the inherited hypernyms of the current token, obtained from WordNet, up to the ``entity'' node (excluded). The lemma of the token and its POS tag are used to perform a WordNet lookup and the first WordNet sense (that usually is the most common one) is chosen. For each of the hypernym synsets the first word is taken and used as a feature. In the case of the example, four features are added:
\begin{itemize}
\singlespacing
\item \texttt{hypernyms\_mcs=body},
\item \texttt{hypernyms\_mcs=social\_group},
\item \texttt{hypernyms\_mcs=group},
\item \texttt{hypernyms\_mcs=abstraction}.
\end{itemize}

\item \textsc{\small{HypernymsDisambiguated}} Same as the \textsc{HypernymsMCS} feature, but instead of selecting the most common sense makes use of the output of the disambiguation step done during preprocessing. If the current token has a Babel synset set, then the corresponding WordNet synset is obtained and the hypernyms are computed from that synset; if the token doesn't have a Babel synset the feature is not set. Although BabelNet provides relations between synsets, and thus hypernyms could have been computed without passing through WordNet, the latter has been preferred for the quality of its manually built taxonomy that, compared to that of BabelNet (that is built automatically) is more accurate and contains less noise.

\item \textsc{\small{VerbHypernymsMCS}} Same as \textsc{HypernymsMCS} but computed on the target verb.

\item \textsc{\small{TokenSimilarWords}} Top 50 similar words to the lemma of the current token. The words are obtained using the DISCO Java library \cite{disco} and a word2vec \cite{word2vec} model computed on the English Wikipedia.

\item \textsc{\small{VerbSimilarWords}} Same as the previous feature, computed for the verb.

\item \textsc{\small{TokenMostSimilarLabels}} The most similar semantic classes to the current token. Exploiting the compositionality of word2vec vectors, a vector representing each semantic class is computed as the average of the vectors of the (lowercased and lemmatized) words that compose the name of the class. For example, the vector for the \emph{Abstract Entity} class is the average of the vector for ``abstract'' and ``entity''. These vectors are precomputed for each semantic class, and then, given a token, its cosine similarity with every class is computed: the top 10 most similar classes are used as features.
\end{enumerate}

\noindent Real-valued features:
\begin{enumerate}
\setcounter{enumi}{64}
\item \textsc{\small{VerbPreps}}
This feature is an indicator of how likely it is that the target verb takes a prepositional argument and how related the verb is to each preposition. To compute this, a list of all the words tagged \emph{advprep} is gathered from the training and test set (see Appendix A for the complete list of words): most of these words are prepositions but some are not; nonetheless, I will refer to them as prepositions in this section because they are to be considered prepositional arguments for the verb. For each word $p$ in this list, and each verb $v$ in the training and test set, the probability that verb $v$ has an adjacent preposition $p$ is computed as:

$$P(v + p) = \frac{count(v + p)}{count(v)}$$

where $count(v + p)$ is the number of times that the verb occurs adjacent to $p$ and $count(v)$ is the occurrences of the verb. These values are computed on a corpus of  around 300000 pages taken from the English version of Wikipedia. This feature is similar to the one used in the verb classification module implemented in \cite{blcunlp}, with some differences. First of all, they compute both the probability that a verb takes a prepositional argument (based on the dependency parse tree) and the probability that the argument is adjacent to the verb: since dependency parsing is slow and requires a lot of time with limited computational power, especially if parsing of a lot of sentences is necessary, I preferred to only compute the probability of the preposition being adjacent to the verb, thus avoiding the need to parse each sentence, since, although this may be less precise, it is very likely that if a preposition is adjacent to a verb it is also one of its dependencies. Secondly, one value is computed for each preposition, instead of computing just one feature grouping all of them: this allows to differentiate verbs that are related to different prepositions from each other. For example, as can be seen from Table \ref{tab-preps}, that shows the value of this feature for the verb ``continue'', this verb is highly correlated with the preposition ``to'' and not, or slightly, correlated with all the other ones. Computing a value for each preposition allows to know which one of them the verb is more related to, instead of just knowing whether the verb is  likely to take an unspecified prepositional argument.

\begin{table}[]
\centering

\begin{tabular}{@{}lllllllll@{}}
\toprule
to  &   in &    on  &   into & through & from  &  over & how & ...\\ \midrule
0.451 & 0.031 & 0.023 & 0.011 & 0.010 & 0.003 & 0.002 & 0.001 & ...\\
\bottomrule
\end{tabular}
\caption{Value of the \textsc{VerbPreps} feature for the verb ``continue''. It can be seen that the verb is highly correlated with the preposition ``to'' and slightly correlated with all the other ones.}
\label{tab-preps}
\end{table}

\item \textsc{\small{VerbDistance}} The distance between the token and the target verb, measured in number of tokens.
\item \textsc{\small{DepDepthDifference}} The difference in depth in the dependency parse tree between the token and the target verb.
\item \textsc{\small{ConDepthDifference}} The difference in depth in the constituency parse tree between the token and the target verb.
\item \textsc{\small{DepPathToVerbLength}} The length of the shortest directed path between the token and the target verb in the dependency parse tree.
\item \textsc{\small{ConPathToVerbLength}} The length of the path between the token and the target verb in the constituency parse tree.
\end{enumerate}

\section{B F-scores with respect to class frequency}

\begin{figure}[H]
\begin{center}
    \includegraphics[height=0.8\textheight]{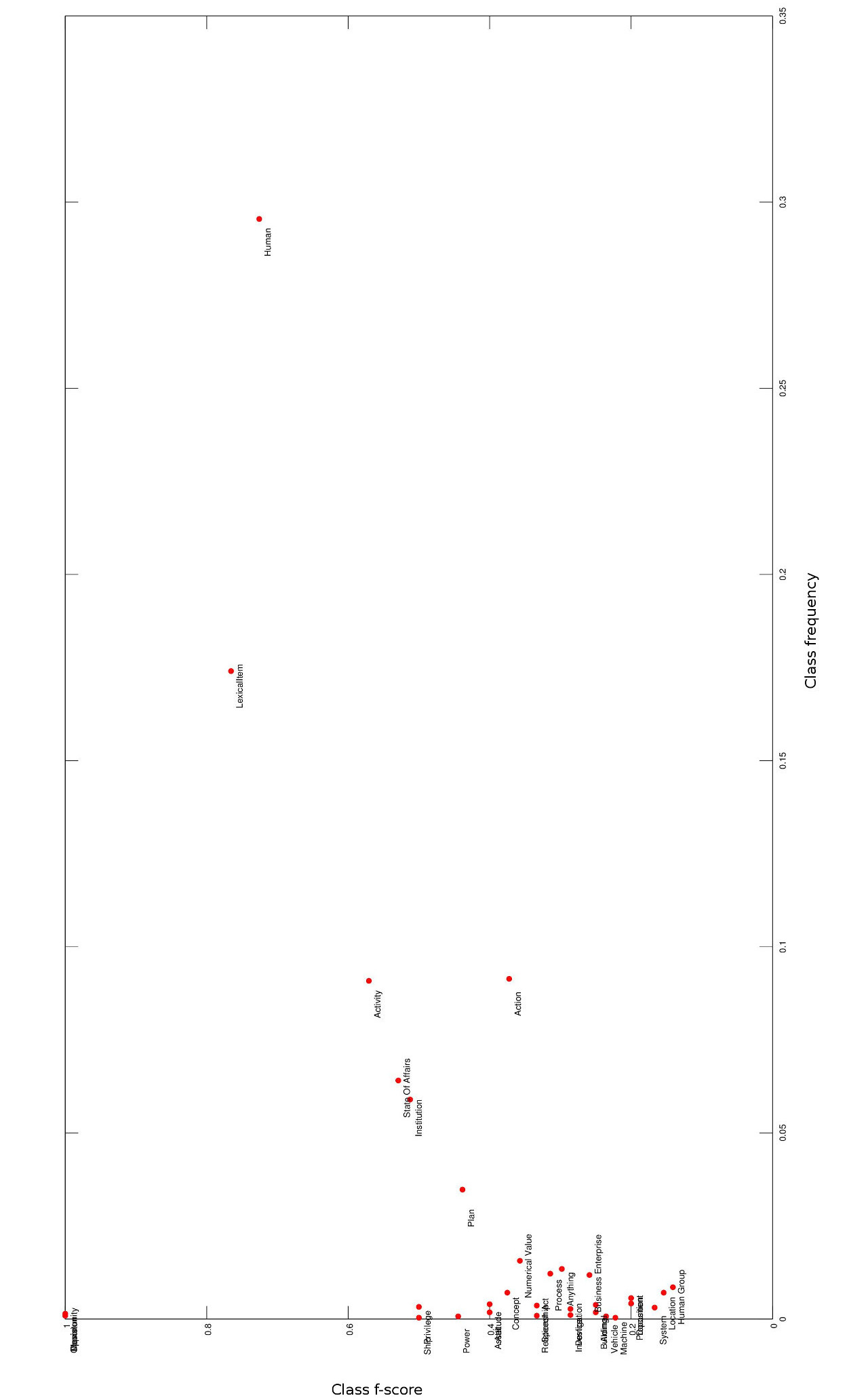}
    \end{center}
    \caption{F-scores for semantic classes with respect to the class frequency in the training set. Classes whose f-score is 0 are not plotted.}
   \label{fig-semantic-fscore}
\end{figure}

\end{document}